\documentclass[11pt]{article}

\usepackage[margin=1in]{geometry}
\usepackage{authblk}
\usepackage[colorlinks,urlcolor=blue,linkcolor=blue,citecolor=blue]{hyperref}
\expandafter\def\expandafter\UrlBreaks\expandafter{\UrlBreaks\do\/\do\*\do\-\do\~\do\'\do\"\do\-}
\usepackage{color}

\usepackage{url}
\usepackage{booktabs}
\usepackage{float}
\usepackage{listings}
\begin{document}

\title{Executable Archaeology of Programs That Model Programming: Reanimating the Heuristic Compiler (1961) and KBEmacs (1984)}

\author[1]{Jeff Shrager}
\author[2]{Kent M.~Pitman}
\author[3]{Mia Golec}
\affil[1]{jshrager@gmail.com, \url{https://shrager.org}}
\affil[2]{kmp@nhplace.com, \url{https://nhplace.com/kent}}
\affil[3]{mgolec@andrew.cmu.edu, Carnegie Mellon University, Pittsburgh, PA, USA}
\date{}

\maketitle

\begin{abstract}

Early AI programs were often also theories of how people think, and
programs that wrote programs were, in part, theories of how people
program. Here we report on the reanimation of two such programs that
offer two seemingly different theories of programming and were developed for
quite different purposes. The first, Herbert Simon's Heuristic
Compiler (1961), was an experimental model of programming as
problem-solving, applying the means-ends analysis methods of the
General Problem Solver to investigate the nature of the programming
task. The second, KBEmacs (1984), was a working prototype of an
intelligent programming assistant developed by MIT's Programmer's
Apprentice project, built on the view that expert programmers
recognize and combine familiar patterns, or ``cliches.''  It let a
programmer build up programs by combining cliches through a stylized
English command language. We approach both through the methodology of
Executable Archaeology, the study of historical computational systems
by running their original code, a process we call ``reanimation.'' Our
reanimation of the Heuristic Compiler reproduces the compiled code
appearing in Simon's 1963 paper, and our reanimation of KBEmacs
reproduces the SQRT sequence of its December 1984 self-test.
Examined side by side, the two theories turn out to have much in
common: the Heuristic Compiler also works by adapting existing
routines, and Simon later argued that search alone requires a large
body of programming knowledge. To
highlight the reflexivity of this project, we deliberately used a
modern AI-assisted programming tool, Anthropic's Claude Code (Opus
5.5), to carry out much of the detailed reanimation work, building on
a significant base of existing software. The
explicit, executable theories of programming that both systems
embodied are largely absent from today's far more capable AI coding
assistants.

\end{abstract}


\section{Introduction}
\label{sec:intro}

Modern programmers increasingly work alongside AI-based tools built on
large language models (LLMs) and trained on enormous corpora of source
code and other programming-related materials. Although the scale and
fluency of the modern tools is truly novel, the aspiration of a
program that takes part in the work of programming is nearly as old as
programming itself. In the 1950s what we now call ``compilation'' was
referred to as ``automatic
programming''~\cite{backus1957fortran,richwaters1988myths}. And within
a few years of the founding of artificial intelligence, researchers
were asking whether the methods of AI could be turned on the task of
programming itself~\cite{simon1963jacm,balzer1973}.

Programming has also long been studied as a human activity. Weinberg's
\textit{The Psychology of Computer Programming}~\cite{weinberg1971}
treated programming as a behavior to be observed, and by the early
1980s Soloway and others theorized that expert programmers usually
work from stereotyped ``programming plans'' -- reusable code schemas
-- rather than from first principles~\cite{soloway1984}. Similarly,
most early AI either embodied formal theories of human
cognition (e.g.,~\cite{newelletat1958, newell1972hps}), or depended
upon intuitive theories of human cognition. In that framing, a program
that writes programs, or helps a programmer do so, is at least an
intuitive theory of programming, but one that could also be helpful to
a human programmer.

This paper describes the reanimation of two historical systems that
offer two such theories, though they were built for different ends.
The first is Herbert A.~Simon's \emph{Heuristic Compiler}, written in
IPL-V in 1961 and described in a RAND report that
year~\cite{simonP2349} and in the \textit{Journal of the ACM} in
1963~\cite{simon1963jacm}. Simon treated programming as a `problem' in
the sense in which that term was used in GPS, the General Problem
Solver of Newell, Shaw, and Simon~\cite{newell1959gps}: Given a
description of what a routine should do, Simon's Heuristic Compiler
searched for differences between the routine as it stood and the
desired routine, and applied operators to reduce those differences.

The second system is \emph{KBEmacs}, the Knowledge-Based Editor (in
Emacs). Built at MIT's Artificial Intelligence Laboratory, KBEmacs was
the culminating demonstration of the Programmer's Apprentice
project~\cite{waters1985tr,waters1986aimag,richwaters1987overview}.
The user of KBEmacs guided the computer by describing the desired
program using stylized English and mentioning specific
``cliches.'' KBEmacs looked up the proto-code underlying the indicated
cliches, supplied the details, and wrote the code by combining the
cliche proto-code into the program under construction. This process
was quite similar to how psychologists studying programming, such as
Soloway (e.g.,~\cite{soloway1984}) theorized human programmers
reasoned (although Soloway used the term ``programming plan'' instead
of ``cliche'').

Although the Heuristic Compiler and KBEmacs were separated by a
quarter of a century, by programming language (IPL-V vs.\ Zetalisp),
and by research tradition (the study of human cognition vs.\ applied
AI), they stake out a great deal of the conceptual territory that
today's AI coding tools occupy: How a programming task is specified,
where the knowledge and skills needed to carry it out come from and
where they reside, how work is divided between human and machine, and
how the system checks what it has done.

We approach the study of these artifacts as exercises in
\emph{Executable Archaeology} employing that field's core methodology,
``reanimation'': studying historical programs by running them on
emulation stacks from code that is as close as possible to the
original~\cite{lane2025eliza,moewsshrager2026}. The Heuristic Compiler
was transcribed from a 1961 line-printer listing recovered from
Simon's papers at CMU, and run on the Common Lisp IPL-V interpreter
built earlier to run the Logic Theorist~\cite{shrager2026}. KBEmacs
was transcribed from a four-volume January 1985 printout and run under
a Zetalisp compatibility layer on top of SBCL, a modern Common
Lisp. Both reanimations were themselves carried out largely by an AI
coding assistant: Anthropic's Claude~\cite{claude}, working in its
command-line form, Claude Code. (The division of labor is described in
``Author Contributions and Use of AI.'')

\section{Historical Background}
\label{sec:history}

\subsection{The Heuristic Compiler}

The goal of Newell, Shaw, and Simon's work at RAND and Carnegie Tech
in the 1950s was to instantiate theories of human thought as computer
programs -- what we (and many others) call ``computational models of
cognition,'' or merely ``models.'' Their first model, developed in 1955--56, was
the Logic Theorist (LT). LT searched for proofs of theorems
in \textit{Principia
  Mathematica}~\cite{LT1957,moewsshrager2026}. Later, they created the
``General Problem Solver'' (1959), which generalized the method applied
in LT into a theory of ``means-ends analysis,'' theorizing that this
is a core cognitive skill~\cite{newell1959gps}. Simon's 1963 summary
of means-ends analysis is worth quoting in full, because the Heuristic
Compiler implements it almost literally:

\begin{quote}
Given the present and desired objects, find a difference between them.
Next, find an operator relevant to the difference; determine if the
operator can be applied to the present object. If so, apply it. (If
not, describe the objects to which it would apply and transform the
present object into an object of that kind -- a new ``desired
object.'') Take the new object thus obtained as the present object and
repeat the process.~\cite[p.~494]{simon1963jacm}
\end{quote}

A proof and a program are not as different as it might seem. A proof
is essentially a program that outputs a theorem, given inputs that are
axioms and prior proved theorems. In this sense (Simon continues), ``a
theorem-proving system can be regarded \ldots\ as a program-writing
system. \ldots\ if we can formulate a programming goal as a difference
between a present and a desired object, we can presumably use the same
processes \ldots\ to generate a
program''~\cite[p.~495]{simon1963jacm}. The same observation, in a
deductive rather than heuristic setting, later became the foundation
of program synthesis from theorem
proving~\cite{green1969,mannawaldinger1971}, and of program verification by
theorem proving~\cite{boyermoore1979}.

Simon's goal was \emph{not} to build an automatic programming tool:
``no claims will be made for the heuristic programming procedures
described here as practical approaches to the construction of
compilers. Their interest lies in what they teach us about the nature
of the programming task''~\cite[p.~493]{simon1963jacm}.\footnote{This
method is called, appropriately, ``task analysis'' in
\cite{newell1972hps}} Two decades later, in a 1984 keynote to the
International Conference on Software Engineering, Simon described
programming by heuristic search from a state description, finding ``a
linking series of operators'' each yielding ``a situation that is, you
hope, a little closer to the final programming goal,'' but judged the
systems built this way to be ``almost toy
systems''~\cite[p.~728]{simon1986}.

\subsection{The Programmer's Apprentice}

The Programmer's Apprentice project began at MIT in the mid-1970s,
with Charles Rich and Howard Shrobe's \textit{Initial Report on a LISP
  Programmer's Apprentice}~\cite{richshrobe1976}. These researchers
were realistic about the potential capabilities of such an AI-based
programming tool at that time: ``Recognizing that it will be a long
time before it is possible to fully duplicate an expert programmer's
abilities, the project seeks to develop an intelligent assistant
system \ldots\ which will help a programmer in various phases of the
programming task''~\cite[p.~47]{waters1986aimag}.

Over the following decade the project developed three central ideas:
\emph{Cliches} are standard computational methods, such as a linear
search, a report generator, or successive approximation, each with
\emph{roles} that vary from use to use and a fixed surrounding
structure. \emph{Plans} are a language-independent representation of
programs as hierarchical flowcharts with explicit data and control
flow, which Rich formalized in a Plan
Calculus~\cite{rich1981,richwaters1987overview}. The \emph{assistant
approach} was a division of labor wherein the programmer makes design
decisions and the apprentice handles the routine details.  Waters's
dissertation developed the analysis of programs into
plans~\cite{waters1978}. Note how these ideas align with those of the
psychology of programming at that same time, for example, Soloway's
programming plans~\cite{soloway1984}. Indeed, the project described
its ultimate goal as tools ``based on theories of how today's expert
programmers analyze, synthesize, modify, explain, and verify their
programs''~\cite{pitman1983wp244}.

KBEmacs was the realization of these ideas in a working editor. It
ran in Zetalisp, the MIT Lisp Machine implementation language, inside
the Zmacs editor of the MIT Lisp machines and their
descendants~\cite{pitman1983wp244}. KBEmacs could create and modify moderately complex
programs, within its range of available cliches -- in Lisp and Ada,
with a mock-up of a PL/I interface~\cite{pitman1983wp244}. Its best-known demonstration builds an Ada report generator,
\texttt{UNIT-REPAIR-REPORT}, from five commands such as ``Define a
simple-report procedure UNIT-REPAIR-REPORT'' and ``Fill the enumerator
with a chain-enumeration of UNITS and
REPAIRS''~\cite[p.~48]{waters1986aimag}, \cite{waters1985tse}.

KBEmacs was admittedly a research prototype. Waters wrote that it knew
``only a few dozen cliches'' where a practical tool ``would need to
know many hundreds,'' and that, its ideas having been used ``for
essentially all [they are] worth,'' KBEmacs ``has been
mothballed''~\cite[pp.~54--55]{waters1986aimag}. The project continued
with work on the Plan Calculus, on automated reasoning, and on
requirements and design apprentices, culminating in the 1990 book
\textit{The Programmer's Apprentice} by Rich and
Waters~\cite{richwaters1990book}. 

\section{Reanimating the Heuristic Compiler}
\label{sec:hc}

\subsection{The Listing}

The source used here for the Heuristic Compiler is a line-printer
listing headed \texttt{HEURISTIC CODER 7/16/61}, discovered in the
Herbert A.~Simon Papers at the Carnegie Mellon University
Archives~\cite{simonlisting1961}. This listing is dated shortly after
the RAND report P-2349~\cite{simonP2349}, and is very nearly the
program described in that report. The listing comprises 94 line-printer
pages and about 4,470 lines (cards). We call this ``the deck.'' The
listing also carries various hand markings indicating a handful of
proposed corrections. The deck contains an executive and two
``compilers''\footnote{Simon uses ``compiler'' at two levels. The
Heuristic Compiler as a whole is a compiler; within it, two
``compilers,'' one for state descriptions and one for functional
descriptions, serve as the operators of its means-ends loop.}; a
natural-language front end (a character scanner, dictionary, and
parser) intended to let the compiler accept English definitions; an
IPL-V knowledge base; and a renderer that takes the final step, in
which the program's internal representation is printed as IPL-V text.

\subsection{Everything Is a Description List}

The central representational idea of the Heuristic Compiler is that a
\emph{routine} is an object with attributes stored as an IPL-V
description list (analogous to a Lisp property list). Each attribute
is one possible definition of the routine: its IPL name, its
\emph{functional description} (i.e., a process name plus arguments,
such as ``INSERT ( ) AT END OF ( )''), its \emph{state description}
(i.e., which cells it affects, and what each holds before and after),
its flow diagram, and its IPL-V code. Within the IPL code rendered in
this representation, IPL-V instructions are themselves description
lists with attributes for each field. For example, a symbol, such as
\texttt{J3}, is stored as a description list with a region (the
letter ``J'') and a location (the integer 3). The program never handles
IPL code as text. Its compilers build and transform these structures,
and the renderer prints them out as IPL-V text only at the end.

The Heuristic Compiler's executive implements a means-ends
problem-solving loop whose \emph{differences} are kinds of definitions
and whose \emph{operators} are called ``compilers.'' The executive
finds the closest definition for the routine, looks up the relevant
compiler in a table of connections, applies it, and tests whether the
routine has improved. If the routine has not improved, the executive
gives up. (There was no search in the AI sense, no handling of failed
paths, and no backtracking.)

The \emph{state description compiler} is the most GPS-like part of the
Heuristic Compiler. It compares the input and output states of each
affected cell, classifies the largest difference (addition, deletion,
replacement, or copying), looks up an operator relevant to it, binds
the operator's variables to the actual cells, and then recursively
compiles two subproblems: getting from the original input state to the
required state, and getting from the operator's output state to the
desired output. Its knowledge of IPL-V consists of exactly two
operators, LOAD and REPLACE, each itself a tiny routine with its own
state description.\footnote{The state description compiler in
\cite{simonRM3588} differs from the one we reanimated in representing
states by bare symbols, whereas ours uses described states.}

The \emph{functional description compiler} works by analogy. Given a
routine described as, say, ``INSERT (1) AT END OF THE VALUE OF THE
ATTRIBUTE (0) OF (2),'' it searches the routines already compiled for
one with the same \emph{process}, here ``INSERT (0) AT END OF (1),''
whose code is the single primitive \texttt{J65}. It then adapts that
routine to the new arguments, moving the inputs into working cells,
compiling code to fetch each argument, recursing when an argument is
itself a described computation (a ``determiner''), and wrapping the
result in instructions that save and restore the working cells.

\subsection{Running the Heuristic Compiler}

The recovered deck runs on the Common Lisp IPL-V interpreter
originally written to run the Logic
Theorist~\cite{shrager2026,shragerrepo}. Claude extended the
interpreter with eleven primitives (J-functions) needed by the
Heuristic Compiler but not required by the Logic Theorist, fixed
several others, and taught its card reader the conventions of the 1961
deck. Two things missing from the 1961 listing were taken from the
1963 listing in RM-3588~\cite{simonRM3588}: a start card, and the
routine U126, which prints a routine's descriptive name. With these
changes, the recovered deck reproduces Simon's published examples.

\textbf{Compiling from a functional description.} Given the functional
description of INSERT AT END OF VALUE LIST, the program produces
\texttt{J13 J52 11W2 11W0 J10 11W1 J65 J32 0}, exactly matching the
code in the 1963 paper~\cite[pp.~499--500]{simon1963jacm}
(Figure~\ref{fig:insert}). A print statement apparently left in routine U113
fortuitously reveals the intermediate stages, one argument at a time,
through which Simon's paper walks its readers. In Simon's runs, as in ours, the functional description is
supplied as a hand-coded list structure, not as typed English; the
phrase is printed from it, but the compiler matches processes by
symbol identity. The English front end in the deck has not yet been
reanimated (see ``Next Steps'').

\begin{figure}[H]
\begin{lstlisting}[escapechar=|]
 FIND THE VALUE-OF THE ATTRIBUTE (0) OF (1)              |$\leftarrow$ \textit{\rmfamily an already-compiled routine}|

                                       J10   0

                                       J65   0           |$\leftarrow$ \textit{\rmfamily the code of INSERT (0) AT END OF (1)}|

                                     11W1                |$\leftarrow$ \textit{\rmfamily fetch argument (1), now in W1}|
                                       J65   0

                                     11W2                |$\leftarrow$ \textit{\rmfamily fetch the determiner's arguments}|
                                     11W0                   |\phantom{$\leftarrow$} \textit{\rmfamily and apply its code (J10)}|
                                       J10
                                     11W1
                                       J65   0

 J13
 INSERT (1) AT END OF  THE VALUE-OF THE ATTRIBUTE (0) OF (2)

                               J13     J52               |$\leftarrow$ \textit{\rmfamily preserve working cells W0-W2}|
                                     11W2
                                     11W0
                                       J10
                                     11W1
                                       J65
                                       J32   0           |$\leftarrow$ \textit{\rmfamily restore them and exit}|
\end{lstlisting}
\caption{Output from the reanimated Heuristic Compiler compiling
  INSERT AT END OF VALUE LIST from its functional description. The
  intermediate stages are printed by the program (via a print in
  routine U113 that appears to be a debugging aid left in the
  deck). The final code matches that displayed in Simon's 1963 paper.
  (The italic annotations marked $\leftarrow$ are ours.)}
\label{fig:insert}
\end{figure}

\textbf{Compiling from a state description.} Simon's running example
of the state description compiler is J3, SET SIGNAL MINUS: a routine
that replaces the top symbol of the signal cell H5 with the symbol J3
(``minus''). The state description says that H5 goes from \emph{some
symbol on top of the rest of its pushdown list} to \emph{J3 on top of
the same rest}. The compiler finds a replacement difference in a named
cell; finds REPLACE relevant, with its variable bound to H5, giving
\texttt{20H5}; sets up the prefix subproblem ``get J3 into the
communication cell,'' an addition, for which LOAD is relevant, giving
\texttt{10J3}; finds no suffix subproblem; and assembles \texttt{10J3
  20H5 0}, the code in the paper~\cite[p.~498]{simon1963jacm}.

We noted one exception in the reproduced results: Our recovered deck
did not initially compile J3. The reason, we discovered, is that our
listing's J3 had both a functional and a state description, but the
executive always prefers the functional description on the assumption
that ``it is easier to compile from a functional description than from
a state description''~\cite[p.~502]{simon1963jacm}. As no
already-compiled routine shares J3's process, the functional
description compiler fails; the executive sees no progress, and gives
up without trying the state description. The published paper's J3
result is only produced by our reanimation when the deck is repaired
by \emph{removing} J3's functional description, after which the state
description compiler produces exactly the published code. In the 1963
listing (in \cite{simonRM3588}), the executive is identical, but the
cards that give J3 a functional description are numbered as though
they were later additions, suggesting that the paper's J3 came from an
earlier version in which, like our repair, J3 had only a state
description.

\textbf{Flow diagrams.} The paper's last section handles routines with
branches and loops by decomposing them into \emph{segments} joined by
a flow diagram. Our reanimation reassembles routine J77 (``test
whether there is a symbol equal to 0ACCUMULATOR on list
1ACCUMULATOR'') from its segments and decomposes it again into the
paper's flow diagram.

\section{Reanimating KBEmacs}
\label{sec:kbe}

\subsection{The Listings}

The KBEmacs source that we used for its reanimation is a January 1985
printout comprised of four bound volumes of Zetalisp code, preserved
by Kent~Pitman, a member of the KBEmacs team. The scans are available
at
\url{https://github.com/netsettler/netsettlement/tree/main/kbe}~\cite{pitmanscans}. These
cover the system as of KBE version 358.0x, a shorthand of the time for ``358.0
(experimental)'' (December 1984), although
individual files carry their own dates and versions. Volume~1 opens
with the output of an extremely useful automatic self-test, run in
December 1984 on one of the MIT Lisp machines or their descendants. The self-test exercises six
command sequences and compares the generated code with the expected
code.  This specific run reports that three sequences (HASH, SQRT, and
SUMLIST) succeeded, but that three (BUCKET, DELETE, and MEMQ) failed.
SQRT was a standing example: an early version of the same scenario
appears in a 1983 description of the KBEmacs interface~\cite{pitman1983wp244}.

The 87 files were transcribed (mostly by ChatGPT~\cite{chatgpt}) into
about 34,500 lines of source, available at
\url{https://github.com/jeffshrager/kbemacs}~\cite{kbemacsrepo}.

\subsection{Text and Plans}

A KBEmacs user works in a stylized English command language phrased in
terms of cliches and their roles. The cliche library is written as
code with holes, as in the definition of successive approximation
in Figure~\ref{fig:cliche}.

\begin{figure}[H]
\begin{lstlisting}
(DEFINE-CLICHE SUCCESSIVE-APPROXIMATION
   (PRIMARY-ROLES (INITIAL-APPROXIMATION)
    DESCRIBED-ROLES (INITIAL-APPROXIMATION CRITERION IMPROVEMENT-STEP)
    COMMENT "iteratively improves an approximation until {the criterion} is
             satisfied" ...)
  (LET* ((APPROXIMATION {the input initial-approximation}) )
    (LOOP DO
      (IF ({the criterion} APPROXIMATION) (RETURN APPROXIMATION) )
      (SETQ APPROXIMATION ({the improvement-step} APPROXIMATION) ))))
\end{lstlisting}
\caption{A cliche from KBEmacs's library (\texttt{kbelib.lisp}, 1984).
The braces are KBEmacs read syntax marking \emph{roles}, the parts that
vary from use to use; the surrounding code is fixed.}
\label{fig:cliche}
\end{figure}
Like the Heuristic Compiler, which worked with internal
representations of IPL-V code, KBEmacs does not store its cliches as
text. When the library loads, each definition is run through
KBEmacs's \emph{analyzer}, which turns the cliche code into a plan: a
set of representations describing segments, ports, data flow, control
flow, and roles. In a plan, a \texttt{LOOP} with a \texttt{SETQ}
disappears; what remains is a temporal composition of a sequence of
states. The knowledge-based editor operates on plans, and after every
command the \emph{coder} turns the plan back into program text,
deciding where to nest expressions and where to introduce variables,
preferring names the programmer has already used. Waters called this
two-way translation the \emph{representation shift}, and argued that
it gave plans locality, additivity, and modularity that text
lacks~\cite{waters1986aimag}. Because the user may also simply edit
the text, KBEmacs must keep text and plan consistent in both
directions.

\subsection{Running KBEmacs (headless)}

KBEmacs depends heavily on the Lisp Machine environment of its time,
including Flavors (the Lisp Machine's object system), the Zmacs
editor, Lisp Machine processes, a special LetS loop notation, and
Zetalisp itself, which is similar to, but not exactly the same as,
modern Common Lisp. Emulators survive for the MIT CADR Lisp Machine
(usim) and for Symbolics Open Genera, but no usable environment
compatible with this KBEmacs release was
available to us. Both include Flavors, but Open Genera, the closer of
the two, is not freely available, and it corresponds to a system
about a decade later than the one KBEmacs ran on.
Even with such an environment, running this release would likely have
required adapting it to a different system version. We chose instead
to write the adaptation explicitly, against modern Common Lisp: the
resulting compatibility layer is itself a record of what KBEmacs
assumed about its machine.

The reanimation targeted the \emph{headless} core: the database,
analyzer, cliche library, command language, knowledge-based editor,
and coder, leaving out the editor, windows, actors, and the Ada and PL/I
front ends~\cite{kbemacsrepo}.

The 1985 source, with correction of minor typos and transcription
errors, is loaded into SBCL (\url{https://www.sbcl.org}), a modern
Common Lisp\footnote{Coincidentally, Spice Lisp, which was built at
CMU, where Simon and Newell taught, begat CMUCL, and SBCL is a fork of
CMUCL.}, on top of a Zetalisp compatibility layer that Claude
assembled and debugged. The bulk of that layer derives from g000001's
Common Lisp port of Spice Lisp Flavors~\cite{g000001}. The rest of the
layer supplies missing Zetalisp and Lisp Machine functionality and
accommodates differences in Common Lisp behavior. A small test harness was also created
to replay the self-test sequences from the 1984 code and compare the
results with what was expected.

The SQRT sequence (Figure~\ref{fig:sqrt}) builds a square-root routine
from three cliches in seven commands, and then builds a variant with
an extra parameter in seven more. After the first command that names a
cliche, KBEmacs holds the cliche's structure with its roles still
open as holes. Each further command fills a role with code or with
another cliche, whose own unfilled roles appear as new holes. After
seven commands the plan has 32 segments and the program is complete.

\begin{figure}[H]
\begin{lstlisting}
>Reinitialize database.
>Define a program MYSQRT with a parameter NUM.
>Implement it as a successive-approximation.
>Implement the initial-approximation as [1.] .
>Implement the improvement-step as an average of [APPROXIMATION]
   and [(// NUM APPROXIMATION)] .
>Implement the criterion as an equality-within-epsilon of
   [(* APPROXIMATION APPROXIMATION)] and [NUM] .
>Implement the epsilon as [0.0001] .

(DEFUN MYSQRT (NUM)
  (LET* ((APPROXIMATION 1))
    (LOOP DO
      (IF (< (ABS (- (* APPROXIMATION APPROXIMATION) NUM)) 0.0001)
          (RETURN APPROXIMATION))
      (SETQ APPROXIMATION (// (+ APPROXIMATION (// NUM APPROXIMATION)) 2.0)))))

The program MYSQRT is a successive-approximation. The initial-approximation is 1.
The criterion is an equality-within-epsilon ...
\end{lstlisting}
\caption{The first half of the SQRT sequence from KBEmacs's December
  14, 1984 self-test: the commands, the generated code, and KBEmacs's
  own description of the program, generated from its plan, all as
  recorded in 1984 (the description is abridged). Replayed in 2026,
  the reanimation generates code \texttt{equal} to this code.}
\label{fig:sqrt}
\end{figure}

All fourteen commands are parsed correctly and the generated programs
are \texttt{equal} (in the Lisp sense) to the 1984 generated
code. The generated programs also work: \texttt{(mysqrt 2.0)} returns
1.4142157.

\section{Then and Now: Two Historical Systems and Today's AI Coding Assistants}
\label{sec:thenandnow}

Neither the Heuristic Compiler nor KBEmacs is an ancestor of today's
AI coding assistants in any direct technical sense. The historical
programs are loaded with explicit proto-programs or cliches, whereas
LLMs are numerical networks trained on enormous corpora of code and
related materials such as discussions about programs in blog posts and
forums.\footnote{One could argue that ``training'' and ``loading'' are
not very different in concept as they both build up background
knowledge. However, the loading of cliches and the training of a
neural net are, of course, completely different forms of
representation.}  Also, the Heuristic Compiler and KBEmacs operate on
explicit internal representations of the code, whereas LLMs operate on
the program text and contextual materials such as the programmer's
guidance prompts.\footnote{Certainly an LLM harbours a tacit internal
representation relevant to the structure of the code it is working on,
but this is, as far as anyone can tell, quite unlike the explicit
representations involved in the reasoning carried out by the
historical programs studied here.} Nonetheless, the questions these
two systems were built to explore are recognizably the questions that
shape AI-assisted programming today. But the comparisons that follow
are conceptual continuities, not architectural equivalences.

\textbf{How is a programming task specified?} Simon's two compilers
differ primarily in how the task is described. A state description
says what the machine looks like before and after, which is close to
what we would now call a specification by examples or by pre- and
post-conditions, and close to the role unit tests play in directing a
modern assistant. A functional description is an imperative phrase,
which is close to a natural-language prompt. Simon drew a conclusion
that remains apt: programmers use both, but ``at the higher levels
\ldots\ a language of functions appears generally the more natural and
powerful,'' and ``the mappings from functional descriptions to process
language are simpler and more direct than the mappings from state
descriptions''~\cite[p.~506]{simon1963jacm}. The authors of KBEmacs
chose the functional side and created a vocabulary of named
algorithms, the cliches. Today's assistants accept both kinds of
description, and much of the craft of using them lies in combining
functional prompts with tests that the AI can use to check its work
and self-correct.

\textbf{Where does the knowledge come from?} Both systems identified
reuse of known solutions as the heart of the matter: Simon's
functional compiler works by finding an analogous routine and adapting
it, and the Programmer's Apprentice rests entirely on the claim that
programs are built from background knowledge. That claim was also a
psychological one: it is essentially what Soloway and his colleagues
were finding in human programmers, whose expertise lay in a store of
``programming plans''~\cite{soloway1984}. Yet both were
knowledge-starved by modern standards. The Heuristic Compiler's state
description compiler knows exactly two operators; its functional
description compiler can compile a routine only if a routine with the
same process has already been compiled. KBEmacs knew ``a few dozen''
cliches; Waters estimated that a practical tool would need ``many
hundreds''~\cite[p.~55]{waters1986aimag}. What neither could solve was
acquiring that knowledge at scale. LLMs can be loosely described as
one answer to precisely this bottleneck: in effect, an LLM is a
vast implicit cliche library absorbed from an enormous body of
code and code-related content. This scale and the LLM's flexibility
come at the price of having no explicit, inspectable representation
of what that body of knowledge is.

\textbf{Who does what?} Simon's Heuristic Compiler was not designed as
a practical programming assistant. It was a computational experiment
intended to illuminate the processes involved in programming. KBEmacs,
by contrast, was deliberately conceived as an assistant, leaving
design decisions to the programmer while taking over routine details.
Its interface was designed to be ``non-invasive'' and
``non-prescriptive,'' always ``willing to defer to the
user''~\cite{pitman1983wp244}.
Waters's argument for the apprentice was practical: full automation
was not near, and a semiexpert assistant could be useful at
once. Modern tools have traversed the same axis in a few years, from
completing a line, to writing a function on request, to agents that
plan, edit, and test across a code base with varying degrees of
supervision~\cite{yang2024sweagent,jimenez2024swebench}. The question
of where the programmer's decisions end and the machine's begin,
central to the Programmer's Apprentice, is again a live design
question. Simon himself put it this way in 1984: ``what part of the
responsibility can we invest in the system itself? \ldots\ what part
of the responsibility still has to be exercised by human beings?
\ldots\ how do we organize things so that the intelligent part of the
automatic system can communicate effectively with the human part of
the intelligence system?''~\cite[p.~727]{simon1986}. His answer was to start with
interactive systems and gradually move ``a boundary between human
beings who are thinking about programs and the intelligent component
of a computer which is thinking about the same
program''~\cite[p.~730]{simon1986}.

\textbf{How does the system know it is making progress?} The Heuristic
Compiler's executive is a loop that directly implements GPS's
means-ends analysis: apply a compiler, test whether the routine has
improved, stop if it reaches the goal or has not progressed. For
Simon, this loop was not merely an engineering choice: means-ends
analysis was Newell and Simon's theory of how people solve
problems~\cite{newelletat1958,newell1972hps}, and the Heuristic
Compiler was a test of whether that theory extended to programming.
Modern coding agents run a loop of a similar shape, explicitly coded
into the harnesses that make these tools work so well: reason about
the task, act on the code, run the tests or the compiler, examine the
difference between what happened and what was wanted, and repair what
fails~\cite{yao2023react,yang2024sweagent,schluntz2024agents,claudecodeworks}. It
has not escaped our notice that the harness used for the reanimations
reported here is among those so described. The difference detector is
now an executed test suite rather than a hand-coded comparison of
states, and the operator is a learned model rather than a table of
connections, but the architecture of means-ends iteration is
recognizable. What survives, however, is the architecture, not the
theory: such harnesses are built because the loop works, not as any
claim about how people program. Yet the resemblance goes beyond the
loop; a coding agent of this kind is not a purely neural system: the
language model chooses what to do next and interprets the results, but
the processes of compiling, executing, comparing outputs, tracking
changes, and so on, are done by conventional symbolic programs (albeit
not, for the most part, AIs). Although the heuristics have moved from
a hand-written table of connections or a cliche library to a neural
net, the operators and the difference tests are still ordinary
programs.

\textbf{Representation.} As we have said, the two historical systems
never treated code merely as text. Both the Heuristic Compiler and
KBEmacs reason from complex internal representations of routines and
cliches, rendering code only when it is time to report the final code,
or, in the case of KBEmacs, when each step commanded by the user is
completed. KBEmacs takes additional pains to keep code and plan
consistent in case the user hand-edits the code directly,
re-analyzing the program when the text no longer matches the code it
last produced~\cite{pitman1983wp244}. Waters
reports that ``well over half of all the effort expended on KBEmacs
went into implementing an analyzer and
coder''~\cite[p.~55]{waters1986aimag}. Today's assistants also only
operate on text at the edges of the network, but their core
representation -- an enormous neural net -- is not easily
interpretable as containing any sort of explicit plan or cliche, a
point we shall return to below.

\section{Technical Next Steps}
\label{sec:next}

For the Heuristic Compiler, the undiscovered country is the English
front end: the scanner, dictionary, and parser, and the routine that
converts a parse into a functional description. Simon credited
H.~S.~Kelly with ``linguistic components for the compiler which are
omitted from this report''~\cite[p.~506]{simon1963jacm}; these
routines may be them. Running them requires input sentences, which are
not in the listing.

The KBEmacs compatibility layer was debugged against one sequence of
the 1984 self-test, SQRT, whose fourteen commands build two programs,
\texttt{MYSQRT} and \texttt{MYOTHERSQRT}. The obvious next step is to
run the remaining five sequences. After that, one could work on the
Ada front end (and the PL/I one), which would allow Waters's main
\texttt{UNIT-REPAIR-REPORT} demonstration to be reproduced. Another
possibility would be to hook KBEmacs in a modern Emacs; reproducing
Zmacs would be difficult and useless, but it would be pleasing and
amusing to be able to enter Control-Meta-Cokebottle into a modern
Emacs and have it bring up KBEmacs!

\section{Conclusion}
\label{sec:conc}

Our reanimations of Simon's Heuristic Compiler and MIT's KBEmacs
recover two apparently different attempts to make programming itself a
subject of computation. Neither was simply an early, less capable
version of today's AI coding assistants; both embodied explicit
accounts of how people program. The Heuristic Compiler treated
programming as means-ends problem solving; the Programmer's Apprentice
treated it as the recognition and combination of familiar cliches. The
two accounts are closer than they first appear. The Heuristic
Compiler's functional description compiler worked by adapting an
already-compiled routine, in effect drawing on a small library of
cliches, and Simon himself argued in 1984 that means-ends analysis
alone was not enough: a programming system also needs indexed
operators and a large body of higher-level
strategies~\cite{simon1986}.

Reanimation lets us examine these accounts in a way the publications
alone do not. The recovered programs reproduce Simon's published
examples and KBEmacs's 1984 SQRT sequence, generating code that
matches the historical output (for KBEmacs, code that is
\texttt{equal} to it and that runs). These are
partial recoveries, and reproducing historical behavior does not
establish that the theories behind it are right. It does mean that the
mechanisms described in the literature can once again be executed,
examined, and tested, rather than known only through prose.

There is a further reflexivity here: much of the work of recovering
these explicit accounts of programming was itself done by a modern AI
coding assistant. A system whose programming knowledge is largely
implicit helped restore systems whose programming knowledge was
deliberately made explicit. Something like Simon's means-ends loop
survives in the harness of such a system, and something like KBEmacs's
cliche library may be present, implicitly, in its weights, but neither
is stated in a form that can be reasoned about. For practical
programming, it may be enough that an assistant writes code that works
and passes its tests. For understanding programming as an intellectual
activity, it is not: to understand how a system programs, we must be
able to reason about how it does so, and we cannot reason about what
is not represented. The Heuristic Compiler and KBEmacs stated their
claims about programming in executable form, and Executable
Archaeology lets us examine those claims again.

\section{Code Availability}
\label{sec:code}

The transcriptions, interpreters, compatibility layer, and Hacker's
Guides to both programs are open source, at
\url{https://github.com/jeffshrager/ExecutableArchaeology} and
\url{https://github.com/jeffshrager/kbemacs}~\cite{earepo,kbemacsrepo}. The
repositories contain the original source scans or provide links to
them.

\section*{Author Contributions and Use of AI}

J.~Shrager conceived and directed the project, searched the Simon and
Newell papers in the Carnegie Mellon University Archives with
M.~Golec, wrote the IPL-V interpreter on which the Heuristic Compiler
runs, checked the results against the historical record, and directed
the revision of the paper. K.~M.~Pitman, a member of the original KBEmacs team,
provided the scans of the January 1985 KBEmacs listings and guidance
on the Zetalisp compatibility layer required to run it. M.~Golec
worked with Shrager over an extended period in the Simon and Newell
archives, and selected and scanned the Heuristic Compiler listing. All
human authors read the paper and provided edits.

Most of the programming work described here was carried out by AI
systems, under the authors' direction. Most of the transcription of
the listings from the page images was done by ChatGPT
(OpenAI)~\cite{chatgpt}. Claude (Anthropic), running as Claude
Code~\cite{claude} with the Claude Opus~5.5 model, corrected the
transcriptions by running the code and checking the scans; adapted and
extended Shrager's IPL-V interpreter; adapted g000001's Common Lisp
ports of Spice Lisp Flavors and LetS~\cite{g000001}; developed the
Zetalisp compatibility layer for KBEmacs, and ran and debugged both
programs until they reproduced their historical output. Claude also
wrote the first draft of this paper. The paper was substantially
revised by and under the authors' direction; ChatGPT also checked the
text and citations. The authors verified the claims against the
listings, the program runs, and the historical sources, and take full
responsibility for the final text.

These two reanimations required about 1,500 calls to the Claude Opus
5.5 model, about 1,600 tool invocations, and about six hours of active
machine time (not counting ChatGPT's extraction of the KBEmacs
listings from the scans).

\section*{Acknowledgments}

We thank Emily Davis of Carnegie Mellon University, who supported Mia
Golec's search of the Herbert A.~Simon Papers in Carnegie Mellon
University's Archives, resulting in locating the Heuristic Coder
listing. We thank g000001 for the Common Lisp port of Spice Lisp
Flavors and LetS (\url{https://github.com/g000001/spice-flavors},
\url{https://github.com/g000001/LetS})~\cite{g000001}. RAND Memorandum
RM-3588-PR (\cite{simonRM3588}) was obtained from bitsavers
(\url{http://bitsavers.org/pdf/rand/ipl/}). The Heuristic Coder
listing is copyright CMU archives, and KBEmacs is copyright MIT. These
are reproduced under the notice in those listings.

Finally, we acknowledge Herb Simon, H.~S.~Kelly, Richard Waters,
Charles Rich, and the other members of the Programmer's Apprentice
project, on whose work this paper rests.

\section*{About the Authors}

\noindent\textbf{Jeff Shrager} is a cognitive scientist, computer
scientist, and molecular biologist. In addition to his work on
Executable Archaeology, he is developing CogLean, a Lean-based library
of formal components that can be used to prove claims about cognitive
models. He is currently Chief Scientist of Bennu Climate, Inc.

\noindent\textbf{Kent M.~Pitman} has been involved in design and
development of innovative technologies to assist with a broad range of
topics, such as hypertext, symbolic algebra, programming languages,
document formatting, and web services---to name just a few. He blogs
about social justice, writes poetry about Climate Change, and has made
a history curation project of computer-related artifacts dredged up
from his basement.

\noindent\textbf{Mia Golec} is an undergraduate student at
Carnegie Mellon University studying Ethics, History, and Public Policy
and Decision Science. She works as a research and project assistant
with the University Archives, supporting external researchers and
various archival projects.

\end{document}